\documentclass[11pt]{article}
\usepackage[left=3cm,right=3cm,top=2cm,bottom=2cm]{geometry} 
\usepackage{amsmath} 
\usepackage{amsfonts}
\usepackage{graphicx}
\usepackage[colorlinks=true,linkcolor=blue]{hyperref}
\usepackage{algorithm}
\usepackage[noend]{algorithmic}
\usepackage{algorithmic}
\usepackage{pgfplots}
\usepackage[makeroom]{cancel}
\usepackage{amsthm,amssymb}
\usepackage{mathtools}
\usepackage{stmaryrd}
\usepackage{multirow}

\newtheorem{theorem}{Theorem}
\newtheorem{lemma}{Lemma}
\newcommand{\on}{\operatorname}
\newcommand{\fromto}{\longrightarrow}
\newcommand{\cX}{\mathcal{X}}
\newcommand{\cQ}{\mathcal{Q}}
\newcommand{\cR}{\mathcal{R}}
\newcommand{\bx}{\boldsymbol{x}}

\newcommand{\bz}{\boldsymbol{z}}
\renewcommand{\vec}[1]{\boldsymbol{#1}}

\begin{document}	
	\title{\textbf{Analogy-Based Preference Learning with Kernels}}
	\date{}
	\author{Mohsen Ahmadi Fahandar, Eyke H{\"u}llermeier\\
		Department of Computer Science, Paderborn University \\
		Pohlweg 49-51, 33098 Paderborn, Germany\\
		\texttt{ahmadim@mail.upb.de, eyke@upb.de} \\}
	\maketitle
	
	\begin{abstract}
		
		Building on a specific formalization of analogical relationships of the form ``A relates to B as C relates to D'', we establish a connection between two important subfields of artificial intelligence, namely analogical reasoning and kernel-based machine learning. More specifically, we show that so-called \emph{analogical proportions} are closely connected to kernel functions on pairs of objects. Based on this result, we introduce the \emph{analogy kernel}, which can be seen as a measure of how strongly four objects are in analogical relationship. As an application, we consider the problem of object ranking in the realm of preference learning, for which we develop a new method based on support vector machines trained with the analogy kernel. Our first experimental results for data sets from different domains (sports, education, tourism, etc.) are promising and suggest that our approach is competitive to state-of-the-art algorithms in terms of predictive accuracy.
		
	\end{abstract}

	\section{Introduction}
	
	In this paper, we establish a connection between analogical reasoning and kernel-based machine learning, which are two important subfields of artificial intelligence.  Essentially, this becomes possible thanks to the observation that a specific formalization of analogical relationships, so-called \emph{analogical proportions} \cite{Miclet2009,prade17}, defines a kernel function on pairs of objects. This relationship is established by means of generalized (fuzzy) equivalence relations as a bridging concept.

	Analogical reasoning has a long tradition in artificial intelligence research, and various attempts at formalizing analogy-based inference can be found in the literature. In this regard, the aforementioned concept of analogical proportion is an especially appealing approach, which has already been used successfully in different problem domains, including classification \cite{bounhas14}, recommendation \cite{hug16}, preference completion \cite{pirlot16}, decision making \cite{billingsley17}, and solving IQ tests \cite{beltran16}.
	
	In spite of its popularity in AI in general, analogical reasoning has not been considered very much in machine learning so far. Yet, analogical proportions have recently been used in the context of preference learning \cite{ahmadi_huellermeier_aaai18}, a branch of machine learning that has received increasing attention in recent years \cite{mpub218}. Roughly speaking, the goal in preference learning is to induce preference models from observational (or experimental) data that reveal information about the preferences of an individual or a group of individuals in a direct or indirect way; the latter typically serve the purpose of predictive modeling, i.e., they are then used to predict the preferences in a new situation. 

	Frequently, the predicted preference relation is required to form a total order, in which case we also speak of a \emph{ranking problem}. In fact, among the problems in the realm of preference learning, the task of ``learning to rank'' has probably received the most attention in the literature so far, and a number of different ranking problems have already been introduced. Based on the type of training data and the required predictions, F\"urnkranz and H\"ullermeier \cite{mpub218} distinguish between the problems of object ranking \cite{Cohen99,kami_as10}, label ranking \cite{Har-Peled2002,Cheng2009,Vembu2011}, and instance ranking \cite{mpub191}.
	
	Building on \cite{ahmadi_huellermeier_aaai18}, the focus of this paper is on the problem of object ranking. Given training data in the form of a set of exemplary rankings of subsets of objects, the goal in object ranking is to learn a ranking function that is able to predict the ranking of any new set of objects. Our contribution is a novel approach to this problem, namely a kernel-based implementation of analog-based object ranking.

	The rest of the paper is organized as follows. In the next section, we recall the setting of object ranking and formalize the corresponding learning problem. Section~3 outlines existing methods for the object ranking task, followed by Section~4 in which the connection between analogical reasoning and kernel-based learning is established. In Section~5, we introduce kernel-based analogical reasoning for the object ranking problem. Finally, we present an experimental evaluation of this approach in Section~6, prior to concluding the paper with a summary and an outline of future work.

	\section{Problem Formulation}
	Consider a reference set of objects, items, or choice alternatives $\mathcal{X}$, and assume each item $\boldsymbol{x} \in \mathcal{X}$ to be described in terms of a feature vector; thus, an item is a vector $\boldsymbol{x}  = (x_1, \ldots , x_d) \in \mathbb{R}^d$ and $\mathcal{X} \subseteq \mathbb{R}^d$. 
	The goal in object ranking is to learn a \emph{ranking function} $\rho$ that accepts any (query) subset 
	$$
	Q = \{ \boldsymbol{x}_1, \ldots , \boldsymbol{x}_n \} \subseteq \mathcal{X}
	$$
	of $n = |Q|$ items as input. As output, the function produces a ranking $\pi \in \mathbb{S}_n$ of these items, where $\mathbb{S}_n$ denotes the set of all permutations of length $n$, i.e., all mappings $[n] \fromto [n]$ (symmetric group of order $n$); $\pi$ represents the total order
	\begin{equation}\label{eq:r}
	\boldsymbol{x}_{\pi^{-1}(1)} \succ \boldsymbol{x}_{\pi^{-1}(2)} \succ \ldots \succ \boldsymbol{x}_{\pi^{-1}(n)} \enspace ,
	\end{equation}
	i.e., $\pi^{-1}(k)$ is the index of the item on position $k$, while $\pi(k)$ is the position of the $k$th item $\boldsymbol{x}_k$ ($\pi$ is often called a \emph{ranking} and $\pi^{-1}$ an \emph{ordering}). Formally, a ranking function is thus a mapping
	\begin{equation}\label{eq:map}
	\rho: \, \cQ \fromto \cR \enspace ,
	\end{equation}
	where $\cQ = 2^\mathcal{X} \setminus \emptyset$ is the \emph{query space} and $\cR = \bigcup_{n \in \mathbb{N}} \mathbb{S}_n$ the \emph{ranking space}. The order relation ``$\succ$'' is typically (though not necessarily) interpreted in terms of preferences, i.e., $\boldsymbol{x} \succ \boldsymbol{y}$ suggests that $\boldsymbol{x}$ is preferred to $\boldsymbol{y}$. 
	A ranking function $\rho$ is learned on a set of training data that consists of a set of rankings 
	\begin{equation}\label{eq:td}
	\mathcal{D} =  \big\{ (Q_1, \pi_1) , \ldots , (Q_M, \pi_M) \big\} \, , 
	\end{equation}
	where each ranking $\pi_\ell$ defines a total order of the set of objects $Q_\ell$. Once a ranking function has been learned, it can be used for making predictions for new query sets $Q$. Such predictions are evaluated in terms of a suitable loss function or performance metric. A common choice is the (normalized) \emph{ranking loss}, which counts the number of inversions between two rankings $\pi$ and $\pi'$:
	\begin{equation*} \label{eq:rankloss}
	d_{RL}(\pi, \pi') = 
	\frac{
		\sum_{1 \leq i , j \leq n}  \llbracket {\pi(i) < \pi(j)} \rrbracket \llbracket {\pi'(i) > \pi'(j)} \rrbracket }{n(n-1)/2} \, ,
	\end{equation*}
	where $\llbracket \cdot \rrbracket$ is the indicator function.
	The ranking function (\ref{eq:map}) sought in object ranking is a complex mapping from the query to the ranking space. An important question, therefore, is how to represent a ``ranking-valued'' function of that kind, and how it can be learned efficiently. 
	
	\section{Previous Work} \label{baselines}
	
	Quite a number of approaches to object ranking and related learning-to-rank problems have already been proposed in the literature. In this section, we give a brief overview of some important state-of-the-art methods that will also be used as baselines in our experiments later on. In this regard, we distinguish between the more traditional utility-based approach, in which a ranking function is represented via an underlying (latent) utility function, and the analogy-based approach recently put forward by Ahmadi Fahandar and H{\"u}llermeier \cite{ahmadi_huellermeier_aaai18}.

	\subsection{Utility-based Approach}
	Most commonly, a ranking function is represented by means of an underlying scoring function 
	$$
	U:\, \mathcal{X} \fromto \mathbb{R} \, , 
	$$
	so that $\boldsymbol{x} \succ \boldsymbol{x}'$ if $U(\boldsymbol{x})> U(\boldsymbol{x}')$. In other words, a ranking-valued function is represented through a real-valued function. Obviously, $U$ can be considered as a kind of utility function, and $U(\boldsymbol{x})$ as a latent utility degree assigned to an item $\boldsymbol{x}$. Seen from this point of view, the goal in object ranking is to learn a latent utility function on a reference set $\mathcal{X}$. Once such function is established, a predicted ordering of items is obtained by sorting them according to their estimated (latent) utilities.

	The representation of a ranking function in terms of a real-valued (utility) function also suggests natural approaches to learning. In particular, two such approaches are prevailing in the literature. 
	
	\subsubsection{Pointwise Approach}
	The first approach focuses on individual objects or ``points'' $\boldsymbol{x}$ in the instance space $\mathcal{X}$, and essentially reduces the ranking problem to a regression problem. Correspondingly, it requires information about target values in the form of preference degrees for individual objects. 
	
	As a representative of this category, we will include \emph{expected rank regression} (ERR) in our experimental study \cite{Kamishima2006,kami_as10}. ERR reduces the problem to standard linear regression. To this end, every training example $(Q, \pi)$ is replaced by a set of data points $(\boldsymbol{x}_i , y_i) \in \mathcal{X} \times \mathbb{R}$. Here, the target $y_i$ assigned to object $\boldsymbol{x}_i \in Q$ is given by 
	$$
	y_i = \frac{\pi(i)}{|Q| + 1} \ .
	$$
	This is justified by taking an expectation over all (complete) rankings of $\mathcal{X}$ (which is assumed to be finite) and assuming a uniform distribution. In spite of this apparently oversimplified assumption, and the questionable transformation of ordinal ranks into numerical scores, ERR has shown quite competitive performance in empirical studies, especially when all rankings in the training data (\ref{eq:td}) are of approximately the same length \cite{tfml}. 
	
	\subsubsection{Pairwise Approach}
	The second idea is to reduce the problem to binary classification; here, the focus
	is on pairs of items, which is the reason why the approach is called the pairwise approach. 
	
	As a first representative of this category, we include a simple reduction to binary classification with linear models. Given a ranking (\ref{eq:r}) as training information, this approach extracts all pairwise preferences $\boldsymbol{x}_{\pi^{-1}(i)} \succ \boldsymbol{x}_{\pi^{-1}(j)}$, $1 \leq i < j \leq n$, and considers these preferences as examples for a binary classification task. This is especially simple if $U$ is a linear function of the form $U(\boldsymbol{x}) = \boldsymbol{w}^\top \boldsymbol{x}$. In this case, $U(\boldsymbol{x}) > U(\boldsymbol{x}')$ if $\boldsymbol{w}^\top \boldsymbol{x} > \boldsymbol{w}^\top \boldsymbol{x}'$, which is equivalent to $\boldsymbol{w}^\top \boldsymbol{z} > 0$ for $\boldsymbol{z} = \boldsymbol{x} - \boldsymbol{x}' \in \mathbb{R}^d$. Thus, from the point of view of binary classification (with a linear threshold model), $\boldsymbol{z}$ can be considered as a positive and $-\boldsymbol{z}$ as a negative example. In principle, any binary classification algorithm can be applied to learn the weight vector $\boldsymbol{w}$ from the set of examples produced in this way. As a representative of this class of methods, we will use support vector machines in our experiments; more specifically, we include Ranking SVM \cite{joachims02} as a state-of-the-art baseline to compare with. 
	
	As a representative nonlinear method, we use a well-known model called RankNet \cite{Burges2005}. It represents a utility function in the form of a feedforward neural network. Given a pair of items $\boldsymbol{x}_i, \boldsymbol{x}_j$, it computes scores $s_i=f(\boldsymbol{x}_i)$ and $s_j=f(\boldsymbol{x}_j)$, and predicts the probability
	\[
	\mathbf{P}( \bx_i \succ \bx_j) = \frac{ 1 }{ 1 + \exp( -\sigma(s_i - s_j) ) } \, . 
	\]
	For an observed preference $\bx_i \succ \bx_j$ or $\bx_j \succ \bx_i$, it adapts the network weights using (stochastic) gradient descent, taking the derivative with respect to the logistic loss as a cost function.

	\subsection{Analogy-based Approach}
	A new approach to object ranking was recently proposed on the basis of analogical reasoning \cite{ahmadi_huellermeier_aaai18}. This approach essentially builds on the following inference pattern: If object $\boldsymbol{a}$ relates to object $\boldsymbol{b}$ as $\boldsymbol{c}$ relates to $\boldsymbol{d}$, and knowing that $\boldsymbol{a}$ is preferred to $\boldsymbol{b}$, we (hypothetically) infer that $\boldsymbol{c}$ is preferred to $\boldsymbol{d}$.
	
	This principle is formalized using the concept of analogical proportion \cite{Miclet2009}. For every quadruple of objects $\boldsymbol{a},\boldsymbol{b},\boldsymbol{c},\boldsymbol{d}$, the latter provides a numerical degree to which these objects are in analogical relation to each other. To this end, such a degree is first determined for each attribute value (feature) separately, and these degrees are then combined into an overall degree of analogy.   
	
	Consider four values $a, b, c, d$ from an attribute domain $\mathbb{X}$. The quadruple $(a,b,c,d)$ is said to be in analogical proportion, denoted by $a:b::c:d$, if ``$a$ relates to $b$ as $c$ relates to $d$''. 
	A bit more formally, the degree of proportion can be expressed as
	\begin{equation}\label{eq:ap}
	E \big( \mathcal{R}(a,b) , \mathcal{R}(c,d) \big) \, ,
	\end{equation}
	where the relation $E$ denotes the ``as'' part of the informal description. $\mathcal{R}$ can be instantiated in different ways, depending on the underlying domain $\mathbb{X}$. 
	
	
	In the case of Boolean variables, where $\mathbb{X} = \{0,1\}$, there are $2^4=16$ instantiations of the pattern $a:b::c:d$, of which only the following 6 satisfy a set of axioms required to hold for analogical proportions: 
	
	\begin{center}
		\begin{tabular}{cccc}
			\hline
			$a$ & $b$ & $c$ & $d$ \\
			\hline
			0 & 0 & 0 & 0 \\
			0 & 0 & 1 & 1 \\
			0 & 1 & 0 & 1 \\
			1 & 0 & 1 & 0 \\
			1 & 1 & 0 & 0 \\
			1 & 1 & 1 & 1 \\
			\hline
		\end{tabular}
	\end{center}
	This formalization captures the idea that $a$ differs from $b$ (in the sense of being ``equally true'', ``more true'', or ``less true'', if the values 0 and 1 are interpreted as truth degrees) exactly as $c$ differs from $d$, and vice versa.
	
	In the numerical case, assuming all attributes to be normalized to the unit interval $[0,1]$, the concept of analogical proportion can be generalized on the basis of generalized logical operators \cite{BOUNHAS201736,dubois16}. In this case, the analogical proportion will become a matter of degree, i.e., a quadruple $(a,b,c,d)$ can be in analogical proportion \emph{to some degree} between 0 and 1. An example of such a proportion, with $\mathcal{R}$ the arithmetic difference $\mathcal{R}(a,b)=a-b$, is the following: 
	\begin{equation}\label{eq:v_a}
	v(a,b,c,d) 
	\begin{cases}
	1- | (a-b) - (c-d)| & \text{if } \on{sign}(a-b) = \on{sign}(c-d)\\
	0              & \text{otherwise.}
	\end{cases}
	\end{equation}
	Note that this formalization indeed generalizes the Boolean case (where $a,b,c,d \in \{0,1 \}$).

	To extend analogical proportions from individual values to complete feature vectors,  the individual degrees of proportion can be combined using any suitable aggregation function, for example the arithmetic mean:
	\begin{equation*}\label{eq:agg}
	v(\boldsymbol{a}, \boldsymbol{b} , \boldsymbol{c} , \boldsymbol{d}) = \frac{1}{d} \sum_{i=1}^d 
	v(a_i , b_i , c_i , d_i)  \, .
	\end{equation*}
	With a measure of analogical proportion at hand, the object ranking task is tackled as follows: Consider any pair of query objects $\boldsymbol{x}_i , \boldsymbol{x}_j \in Q$. Every preference $\boldsymbol{z} \succ \boldsymbol{z}'$ observed in the training data $\mathcal{D}$, such that $(\boldsymbol{z}, \boldsymbol{z}', \boldsymbol{x}_i , \boldsymbol{x}_j)$ are in analogical proportion, suggests that $\boldsymbol{x}_i \succ \boldsymbol{x}_j$. This principle is referred as \emph{analogical transfer} of preferences, because the observed preference for $\boldsymbol{z} , \boldsymbol{z}'$ is (hypothetically) transferred to $\boldsymbol{x}_i, \boldsymbol{x}_j$. Accumulating all pieces of evidence that can be collected in favor of $\boldsymbol{x}_i \succ \boldsymbol{x}_j$ and, vice versa, the opposite preference $\boldsymbol{x}_j \succ \boldsymbol{x}_i$, an overall degree $p_{i,j}$ is derived for this pair of objects. The same is done for all other pairs in the query. Eventually, all these degrees are combined into an overall consensus ranking. We refer to \cite{ahmadi_huellermeier_aaai18} for a detailed description of this method, which is called ``\textbf{a}nalogy-\textbf{b}ased \textbf{le}arning to rank'' (able2rank) by the authors. 

	As an aside, note that an analogy-based approach as outlined above appears to be specifically suitable for \emph{transfer learning}. This is mainly because the relation $\mathcal{R}$ is evaluated separately for ``source objects'' $a$ and $b$ on the one side and ``target objects'' $c$ and $d$ on the other side, but never between sources and targets. In principle, one could even think of using different specifications of $\mathcal{R}$ for the source and the target.

	\section{Analogy and Kernels}
	
	The core idea of our proposal is based on the observation that an analogical proportion, by definition, defines a kind of \emph{similarity} between the relation of pairs of objects: According to (\ref{eq:ap}), the analogical proportion $a:b::c:d$ holds if $\mathcal{R}(a,b)$ is similar to $\mathcal{R}(c,d)$. The notion of similarity plays an important role in machine learning in general, and in kernel-based machine learning in particular. In fact, kernel functions can typically be interpreted in terms of similarity. Thus, a kernel-based approach might be a natural way to incorporate analogical reasoning in machine learning. 
	
	More specifically, to establish a connection between kernel-based machine learning and analogical reasoning, we make use of generalized (fuzzy) equivalence relations as a bridging concept. Fuzzy equivalences are weakened forms of standard equivalence relations, and hence capture the notion of similarity. More specifically, a fuzzy equivalence relation $E$ on a set $\mathcal{X}$ is a fuzzy subset of $\mathcal{X} \times \mathcal{X}$, that is, a function $E:\, \mathcal{X}^2 \longrightarrow [0,1]$, which is reflexive, symmetric, and $\top$-transitive:
	\begin{itemize}
		\item $E(x,x) = 1$ for all $x \in \mathcal{X}$,
		\item $E(x,y)=E(y,x)$ for all $x, y \in \mathcal{X}$,
		\item $\top(E(x,y), E(y,z)) \leq E(x,z)$ for all $x, y, z \in \mathcal{X}$,
	\end{itemize}
	where $\top$ is a triangular norm (t-norm), that is, a generalized logical conjunction.
	In our case, the relation $E$ in (\ref{eq:ap}) will play the role of a fuzzy equivalence. The detour via fuzzy equivalences is motivated by the result of Moser \cite{Moser2006}, who proved that certain types of fuzzy equivalence relations satisfy the properties of a kernel function. Before elaborating on this idea in more detail, we briefly recall some basic concepts of kernel-based machine learning as needed for this paper. For a thorough discussion of kernel methods, see for instance \cite{Scholkopf2001,ShaweTaylor2004}.
	
	\subsection{Kernels}

	Let $\mathcal{X}$ be a nonempty set. A function $k: \mathcal{X} \times \mathcal{X} \longrightarrow \mathbb{R}$ is a {\em positive semi-definite kernel} on $\mathcal{X}$ iff it is symmetric, i.e., $k(x,y) = k(y,x)$ for all $x,y \in \mathcal{X}$,  and positive semi-definite, i.e., 
	\[ \sum_{i=1}^n \sum_{j=1}^n c_i c_j k(x_i,x_j) \geq 0 \]
	for arbitrary $n$, arbitrary instances $x_1, \ldots, x_n \in \cX$ and arbitrary $c_1, \ldots, c_n \in \mathbb{R}$.
	Given a kernel $k$ on $\mathcal{X}$, an important theorem by Mercer \cite{Mercer} implies the existence of a (Hilbert) space $\mathcal{H}$ and a map $\phi:\, \mathcal{X} \longrightarrow \mathcal{H}$, such that 
	$$
	k(x,y) = \langle \phi(x) , \phi(y) \rangle 
	$$
	for all $x,y \in \mathcal{X}$. Thus, computing the kernel $k(x,y)$ in the original space $\mathcal{X}$ is equivalent to mapping $x$ and $y$ to $\mathcal{H}$ first, using the \emph{linearization} or \emph{feature map} $\phi$, and combining them in terms of the inner product in that space afterward. This connection between a nonlinear combination of instances in the original space $\mathcal{X}$ and a linear combination in the induced feature space $\mathcal{H}$ provides the basis for the so-called ``kernel trick'', which offers a systematic way to design nonlinear extensions of methods for learning linear models. The kernel trick has been applied to various methods and has given rise to many state-of-the-art machine learning algorithms, including support vector machines, kernel principle component analysis, kernel Fisher discriminant, amongst others \cite{Scholkopf_nips,Scholkopf98}.

	\subsection{Analogical Proportions as Kernels}

	Our focus in this paper is the analogical proportion (\ref{eq:v_a}), which is a map $v:\, [0,1]^4 \longrightarrow [0,1]$. In this case, the relation $\mathcal{R}$ is the simple arithmetic difference $\mathcal{R}(a,b)=a-b$, and the similarity relation $E$ is defined as $E(u, v) = 1- |u-v|$ if both $u,v$ have the same sign and $E(u, v) = 0$ otherwise. As an aside, we note that, strictly speaking, $E$ thus defined is not a fuzzy equivalence relation. This is due to the thresholding in the case where $\text{sign} (a-b) \neq \text{sign} (c-d)$. Without this thresholding, $E$ would be a $\top_{\!\!\L}$-equivalence, where $\top_{\!\!\L}$ is the {\L}ukasievicz t-norm $(\alpha,\beta) \mapsto \max(\alpha+\beta-1,0)$. For modeling analogy, however, setting $E$ to 0 in the case where $b$ deviates positively from $a$ while $d$ deviates negatively from $c$ (or vice versa) appears reasonable. 
	
	We reinterpret $v$ as defined above as a kernel function $k:\, [0,1]^2 \times [0,1]^2  \longrightarrow [0,1]$ on $\mathcal{X} = [0,1]^2$, i.e., a kernel on pairs of pairs of objects, which essentially means equating $k$ with $E$:
	\begin{equation}\label{eq:akernel}
	k(a,b,c,d)  \mapsto 
	1- |(a-b) - (c-d)|
	\end{equation}
	if $\text{sign} (a-b) = \text{sign} (c-d)$ and 0 otherwise. 
	In what follows, we show that the ``analogy kernel'' (\ref{eq:akernel}) does indeed define a proper kernel function. The first property to be fulfilled, namely symmetry, is obvious. Thus, it remains to show that $k$ is also positive semi-definite, which is done in Theorem 2 below. As a preparation, we first recall the following lemma, which is proved by Moser \cite{Moser2006} as part of his Theorem 11.
	
	\begin{lemma} \label{lma1}
		Let $\mu_1, \ldots , \mu_n \in [0,1]$, $n \in \mathbb{N}$, and the matrix $M$ be defined by
		\[
		M^{(n)}_{i,j} = ( 1- | \mu_i - \mu_j | ) \,\, .
		\]
		Then $M$ has non-negative determinant.
	\end{lemma}
	
	\begin{theorem}
		The function $k:\, [-1,1]^2 \longrightarrow [0,1]$ defined as
		\begin{equation*} \label{eq:kernel}
		k(u,v) = 
		\begin{cases}
		1 - |u-v|  & \text{if } \text{sign}(u)=\text{sign}(v), \\
		0,              & \text{otherwise,}
		\end{cases}
		\end{equation*}
		is a valid kernel.
	\end{theorem}
	
	\begin{proof}
		It is easy to see that $k$ is symmetric. Thus, it remains to show that it is positive semi-definite. To this end, it suffices to show that the determinants of all principal minors of every kernel matrix produced by $k$ are non-negative. Thus, consider $\alpha_1, \ldots , \alpha_n \in [-1,1]$, $n \in \mathbb{N}$, and the matrix $K$ defined as
		\begin{equation}
		K^{(n)}_{i,j} = 
		\begin{cases}
		1 - |\alpha_i-\alpha_j| ,& \text{if } \text{sign}(\alpha_i)=\text{sign}(\alpha_j), \\
		0,              & \text{otherwise,}
		\end{cases}
		\end{equation}
		We need to show that 
		\[
		\det \bigg ( K^{(m)}_{i,j} \bigg ) \ge 0 \, ,
		\]
		for all $1 \le m \le n$.
		Thanks to the permutation-invariance of determinants, we can assume (without loss of generality) that the values $\alpha_i$ are sorted in non-increasing order, i.e., $\alpha_1 \geq \alpha_2 \geq \cdots \geq \alpha_n$; in particular, note that the positive $\alpha_i$ will then precede all the negative ones. Thus, the matrix $K$ takes the form of a diagonal block matrix
		\[
		K = \begin{pmatrix}
		A & 0 \\
		0 & B \\
		\end{pmatrix} \, ,
		\]
		in which the submatrix $A$ contains the values of $K$ for which $\alpha_i, \alpha_j \in [0,1]$, and $ B $ contains the values of $K$ where $\alpha_i, \alpha_j$ are negative. According to Lemma (\ref{lma1}), $\det(A) \ge 0$. Moreover, since $1-|u-v| = 1-|(-u)-(-v)|$ for $u,v \in [0,1]$, the same lemma can also be applied to the submatrix $B$, hence $\det(B) \ge 0$. Finally, we can exploit that
		\[
		\det(K) = \det(A) \det(B).
		\]
		Since both matrices $A$ and $B$ have non-negative determinant, it follows that $\det(K) \ge 0$, which completes the proof.
	\end{proof}

	The class of kernel functions is closed under various operations, including addition and multiplication by a positive constant. This allows us to extend the analogy kernel from individual variables to feature vectors using the arithmetic mean as an aggregation function:
	\begin{equation} \label{eq:analogykernel}
	k_A( \boldsymbol{a}, \boldsymbol{b} , \boldsymbol{c}, \boldsymbol{d}) =
	\frac{1}{d} \sum_{i=1}^d
	k(a_i , b_i, c_i, d_i)  \, .
	\end{equation}

	Furthermore, to allow for incorporating a certain degree of non-linearity, we make use of a homogeneous polynomial kernel of degree 2,
	\begin{equation} \label{eq:polykernel}
	k_A'( \boldsymbol{a}, \boldsymbol{b} , \boldsymbol{c}, \boldsymbol{d} ) = \big( k( \boldsymbol{a}, \boldsymbol{b} , \boldsymbol{c}, \boldsymbol{d} ) \big)^2 \, ,
	\end{equation}
	which is again a valid kernel.

	\section{Analogy-Kernel-Based Object Ranking}
	
	Recall that, in the setting of learning to rank, we suppose to be given a set of training data in the form 
	$$
	\mathcal{D} = \big\{ (Q_1, \pi_1) , \ldots , (Q_M, \pi_M) \big\} \ ,
	$$
	where each $\pi_\ell$ defines a ranking of the set of objects $Q_\ell$. If $\boldsymbol{z}_i , \boldsymbol{z}_j \in Q_\ell$ and $\pi_\ell(i) < \pi_\ell(j)$, then $\boldsymbol{z}_i \succ \boldsymbol{z}_j$ has been observed as a preference.
	Our approach to object ranking based on the analogy kernel, AnKer-rank, comprises two main steps:
	\begin{itemize}
		\item First, for each pair of objects $\boldsymbol{x}_i , \boldsymbol{x}_j \in Q$, a degree of preference $p_{i,j} \in [0,1]$ is derived from $\mathcal{D}$. If these degrees are normalized such that $p_{i,j} + p_{j,i} = 1$, they define a reciprocal preference relation
		\begin{equation}\label{eq:prefrel}
		P= \Big( p_{i,j} \Big)_{1 \leq i \ne j \leq n} \, .
		\end{equation}
		\item Second, the preference relation $P$ is turned into a ranking $\pi$ using a suitable ranking procedure. 
	\end{itemize}
	Both steps will be explained in more detail further below. 

	\subsection{Prediction of Pairwise Preferences}
	
	The first step of our proposed approach, prediction of pairwise preferences, is based on a reduction to binary classification. To this end, training data $\mathcal{D}_{bin}$ is constructed as follows. Consider any preference $\boldsymbol{x}_i \succ \boldsymbol{x}_j$ that can be extracted from the original training data $\mathcal{D}$, i.e., from any of the rankings $\pi_m$, $m \in [M]$. Then $\bz_{i,j}=(\bx_i, \bx_j)$ is a positive example for the binary problem (with label $y_{i,j}=+1$), and $\bz_{j,i}=(\bx_j, \bx_i)$ is a negative example (with label $y_{j,i}=-1$). Since these examples essentially carry the same information, we only add one of them to $\mathcal{D}_{bin}$. To keep a balance between positive and negative examples, the choice is simply made by flipping a fair coin. 
	
	Note that, for any pair of instances $(\vec{a}, \vec{b})$ and $(\vec{c}, \vec{d})$ in $\mathcal{D}_{bin}$, the analogy kernel (\ref{eq:analogykernel}) is well-defined,  i.e., $k_A(\vec{a}, \vec{b}, \vec{c}, \vec{d})$ can be computed. Therefore, a binary predictor $h_{bin}$ can be trained on $\mathcal{D}_{bin}$ using any kernel-based classification method. We assume $h_{bin}$ to produce predictions in the unit interval $[0,1]$, which can be achieved, for example, by means of support vector machines with a suitable post-processing such as Platt-scaling \cite{Platt99probabilisticoutputs}.

	Now, consider any pair of objects $\bx_i, \bx_j$ from a new query $Q=\{ \bx_1, \ldots , \bx_n \}$. Again, the analogy kernel can be applied to this pair and any example from $\mathcal{D}_{bin}$, so that a (binary) prediction for the preference between $\bx_i$ and $\bx_j$ can be derived from $h_{bin}$. More specifically, querying this model with $\bz_{i,j}=(\bx_i, \bx_j)$ yields a degree of support $q_{i,j}=h_{bin}(\bz_{i,j})$ in favor of $\bx_i \succ \bx_j$, while querying it with $\bz_{j,i}=(\bx_j, \bx_i)$ yields a degree of support $q_{j,i}=h_{bin}(\bz_{j,i})$ in favor of $\bx_i \succ \bx_j$. As already said, we assume both degrees to be normalized in the range $[0,1]$ and define $p_{i,j} = (1+q_{i,j}-q_{j,i})/2$ as an estimate for the probability of the preference $\bx_i \succ \bx_j$. This estimate constitutes one of the entries in the preference relation (\ref{eq:prefrel}).  

	\subsection{Rank Aggregation}
	
	To turn pairwise preferences into a total order, we make use of a rank aggregation method. More specifically, we apply the Bradley-Terry-Luce (BTL) model, which is well-known in the literature on discrete choice \cite{brad_tr52}. It starts from the parametric model 
	\begin{equation}\label{eq:pmp}
	\mathbf{P}(\boldsymbol{x}_i \succ \boldsymbol{x}_j) = \frac{\theta_i}{\theta_i + \theta_j} \, ,
	\end{equation}
	where $\theta_i, \theta_j \in \mathbb{R}_+$ are parameters representing the (latent) utility $U(\boldsymbol{x}_i)$ and $U(\boldsymbol{x}_j)$ of $\boldsymbol{x}_i$ an $\boldsymbol{x}_j$, respectively. Thus, according to the BTL model, the probability to observe a preference in favor of a choice alternative $\boldsymbol{x}_i$, when being compared to any other alternative, is proportional to $\theta_i$. 
	
	Given the preference relation (\ref{eq:prefrel}), i.e., the entries $p_{i,j}$ informing about the class probability of $\boldsymbol{x}_i \succ \boldsymbol{x}_j$, the parameter $\theta = (\theta_1, \ldots , \theta_n)$ can be estimated by likelihood maximization:
	$$
	\hat{\theta} \in \arg \max_{\theta \in \mathbb{R}^{n} } \prod_{1 \leq i \neq j \leq n}  \left( \dfrac{\theta_{i}}{\theta_{i} + \theta_{j}} \right)^{p_{i,j}}
	$$
	Finally, the predicted ranking $\pi$ is obtained by sorting the items $\boldsymbol{x}_i$ in descending order of their estimated (latent) utilities $\hat{\theta}_i$. 
	
	We note that many other rank aggregation techniques have been proposed in the literature and could principally be used as well; see e.g.\ \cite{pmlr-v70-fahandar17a}. However, since BTL seems to perform very well, we did not consider any other method.
	
	\section{Experiments}
	
	To study the practical performance of our proposed method, we conducted experiments on several real-world data sets, using able2rank, ERR, Ranking SVM (with linear kernel) and RankNet (cf.\ Section \ref{baselines}) as baselines to compare with.
	
	\begin{table*}
		\caption{Properties of data sets.}
		\label{tab:ds}
		\footnotesize
		\centering
		\begin{tabular}{ |c||lccccc|c| } 
			\hline
			data set & domain & \# instances & \# features & numeric & binary & ordinal & name \\
			\hline
			\multirow{4}{*}{Decathlon} & Year 2005 & 100 & 10 & x& -- & -- & D1\\ 
			& Year 2006 & 100 & 10 & x & -- & -- & D2 \\ 
			& Olympic 2016 & 24 & 10 & x & -- & -- & D3 \\ 
			& U-20 World 2016  & 22 & 10 & x & -- & -- & D4 \\ 
			\hline
			\multirow{3}{*}{Bundesliga} & Season 15/16 & 18 & 13 & x& -- & -- & B1\\ 
			& Season 16/17 & 18 & 13 & x & -- & -- & B2 \\ 
			& Mid-Season 16/17 & 18 & 7 & x & -- & -- & B3 \\ 
			\hline
			\multirow{2}{*}{Footballers} 
			& Year 2016 (Streaker) & 50 & 40 & x & x & x & F1 \\ 
			& Year 2017 (Streaker)& 50 & 40 & x & x & x & F2 \\ 
			\hline
			\multirow{3}{*}{FIFA WC} & WC 2014 Brazil & 32 & 7 & x& -- & -- & G1\\ 
			& WC 2018 Russia & 32 & 7 & x & -- & -- & G2 \\ 
			& U-17 WC 2017 India & 22 & 7 & x & -- & -- & G3 \\ 
			\hline
			\multirow{2}{*}{Hotels} & D{\"u}sseldorf & 110 & 28 & x& x & x & H1\\ 
			& Frankfurt & 149 & 28 & x & x & x & H2 \\ 
			\hline
			\multirow{2}{*}{Uni. Rankings} & Year 2015 & 100 & 9 & x& -- & -- & U1\\ 
			& Year 2014 & 100 & 9 & x & -- & -- & U2 \\ 
			\hline
			\multirow{2}{*}{Volleyball WL} & Group 3 & 12 & 15 & x& -- & -- & V1\\ 
			& Group 1 & 12 & 15 & x & -- & -- & V2 \\ 
			\hline
		\end{tabular}
	\end{table*}
	
	\subsection{Data}
	
	We used the same data sets as \cite{ahmadi_huellermeier_aaai18}, which are collected from various domains (e.g., sports, education, tourism) and comprise different types of feature (e.g., numeric, binary, ordinal). Table (\ref{tab:ds}) provides a summary of the characteristics of the data sets. For a detailed description of the data, we refer the reader to the source paper. In addition, we include the ranking of the teams that participated in the men's FIFA world cup 2014 and 2018 (32 instances) as well as under-17 in the year 2017 (22 instances) with respect to ``goals statistics''. This data\footnote{Extracted from FIFA official website: \url{www.fifa.com}} comprises 7 numeric features such as MatchesPlayed, GoalsFor, GoalsScored, etc.
	
	\begin{table*}
		\centering
		\caption{Results in terms of loss $d_{RL}$ (averaged over 20 runs) on the test data.}
		\label{tab:results}
		\hspace*{-2em}
		\begin{tabular}{c|c|ccccc}
			\hline
			$D_{train}$ $\rightarrow$ $D_{test}$ & AnKer-rank & able2rank & ERR & Ranking SVM & RankNet \\ 
			\hline
			D1 $\rightarrow$ D2 & $0.188 \pm 0.049 (5)$ & $0.055 \pm 0.000 (4)$ & $0.053 (3)$ & $0.014 (1)$ & $0.029 \pm 0.005 (2)$  \\
			
			D1 $\rightarrow$ D3 & $0.183 \pm 0.043 (5)$ & $0.072 \pm 0.000 (4)$ & $0.054 (3)$ & $0.040 (2)$ & $0.024 \pm 0.007 (1)$  \\
			
			D1 $\rightarrow$ D4 & $0.187 \pm 0.047 (5)$ & $0.119 \pm 0.002 (4)$ & $0.117 (3)$ & $0.095 (1)$ &  $0.102 \pm 0.009 (2)$  \\
			
			D2 $\rightarrow$ D1 & $0.195 \pm 0.034 (5)$ & $0.090 \pm 0.000 (4)$ & $0.056 (3)$ & $0.015 (1)$  & $0.041 \pm 0.005 (2)$ \\
			
			D2 $\rightarrow$ D3 & $0.102 \pm 0.028 (5)$ & $0.082 \pm 0.002 (4)$ & $0.025 (1)$ & $0.032 (2)$ & $0.032 \pm 0.011 (2)$ \\
			
			D2 $\rightarrow$ D4 & $0.218 \pm 0.040 (5)$ & $0.143 \pm 0.000 (4)$ & $0.126 (3)$ & $0.104 (1)$ &  $0.105 \pm 0.004 (2)$ \\
			
			D3 $\rightarrow$ D1 & $0.133 \pm 0.007 (2)$ & $0.150 \pm 0.000 (4)$ & $0.145 (3)$ & $0.096 (1)$ & $0.226 \pm 0.058 (5)$  \\
			
			D3 $\rightarrow$ D2 & $0.107 \pm 0.007 (3)$ & $0.106 \pm 0.000 (2)$ & $0.109 (4)$ & $0.082 (1)$ & $0.184 \pm 0.023 (5)$  \\
			
			D3 $\rightarrow$ D4 & $0.134 \pm 0.008 (2)$ & $0.144 \pm 0.003 (4)$ & $0.143 (3)$ & $0.126 (1)$ & $0.206 \pm 0.037 (5)$  \\
			
			D4 $\rightarrow$ D1 & $ 0.108 \pm 0.008 (1) $ & $0.156 \pm 0.000 (4)$ & $0.132 (3)$ & $0.119 (2)$ & $0.177 \pm 0.047 (5)$  \\
			
			D4 $\rightarrow$ D2 & $0.115 \pm 0.008 (2)$ & $0.144 \pm 0.000 (5)$ & $  0.105 (1) $ & $0.118 (3)$ & $0.128 \pm 0.014 (4)$  \\
			
			D4 $\rightarrow$ D3 & $0.101 \pm 0.014 (3)$ & $0.099 \pm 0.002 (1)$ & $0.127 (5)$ & $0.101 (3)$  & $0.099 \pm 0.037 (1)$  \\
			\hline
			average ranks &  3.58 & 3.67 &  2.92 & 1.58 & 3.00 \\
			\hline 
			B1 $\rightarrow$ B2 & $  0.018 \pm 0.005 (1) $ & $0.031 \pm 0.006 (2)$ & $0.065 (4)$ & $0.052 (3)$ & $0.104 \pm 0.033 (5)$  \\ 
			
			B1 $\rightarrow$ B3 & $ 0.011 \pm 0.003 (1) $ & $0.013 \pm 0.000 (2)$ & $0.026 (4)$ & $0.020 (3)$ &  $0.056 \pm 0.027 (5)$  \\ 
			
			B2 $\rightarrow$ B1 & $  0.001 \pm 0.002 (1) $ & $0.013 \pm 0.005 (2)$ & $0.118 (5)$ & $0.045 (3)$ & $0.096 \pm 0.022 (4)$  \\
			
			B2 $\rightarrow$ B3 & $  0.000 \pm 0.000 (1) $ & $0.013 \pm 0.000 (2)$ & $0.033 (4)$ & $0.032 (3)$ &  $0.043 \pm 0.019 (5)$  \\ 
			
			B3 $\rightarrow$ B1 & $0.000 \pm 0.000 (1)$ & $0.000 \pm 0.000 (1)$ & $0.007 (3)$ & $0.007 (3)$ & $0.053 \pm 0.024 (5)$  \\
			
			B3 $\rightarrow$ B2 & $  0.000 \pm 0.001 (1) $ & $0.010 \pm 0.003 (3)$ & $0.007 (2)$ & $0.092 (4)$ &  $0.092 \pm 0.024 (4)$  \\ 
			\hline
			average ranks &  1.00 & 2.00 &  3.67 & 3.17 & 4.67 \\
			\hline 
			F1 $\rightarrow$ F2 & $0.183 \pm 0.027 (4)$ & $0.139 \pm 0.001 (1)$ & $0.314 (5)$ & $  0.166 (2) $ &  $0.173 \pm 0.006 (3)$  \\
			
			F2 $\rightarrow$ F1 & $0.155 \pm 0.003 (2)$ & $  0.152 \pm 0.001 (1) $ & $0.293 (5)$ & $0.183 (4)$ & $0.163 \pm 0.009 (3)$  \\
			\hline
			average ranks &  3.00 & 1.00 &  5.00 & 3.00 & 3.00 \\
			\hline 
			G1 $\rightarrow$ G2 & $ 0.040 \pm 0.006 (1) $ & $0.061 \pm 0.003 (3)$ & $0.102 (5)$ & $0.053 (2)$ & $0.085 \pm 0.009 (4)$  \\
			
			G1 $\rightarrow$ G3 & $  0.001 \pm 0.003 (1) $ & $0.012 \pm 0.002 (3)$ & $0.056 (5)$ & $0.004 (2)$ &  $0.044 \pm 0.010 (4)$  \\
			
			G2 $\rightarrow$ G1 & $0.030 \pm 0.001 (2)$ & $ 0.026 \pm 0.002 (1) $ & $0.100 (5)$ & $0.037 (3)$ &  $0.045 \pm 0.008 (4)$  \\
			
			G2 $\rightarrow$ G3 & $0.022 \pm 0.001 (1)$ & $0.025 \pm 0.002 (3)$ & $0.065 (5)$ & $0.022 (1)$ &  $0.047 \pm 0.014 (4)$  \\
			
			G3 $\rightarrow$ G1 & $0.034 \pm 0.008 (3)$ & $0.042 \pm 0.005 (4)$ & $0.029 (2)$ & $0.023 (1)$ &  $0.118 \pm 0.012 (5)$  \\
			
			G3 $\rightarrow$ G2 & $0.098 \pm 0.019 (3)$ & $0.106 \pm 0.004 (4)$ & $0.088 (2)$ & $  0.052 (1) $ &  $0.168 \pm 0.021 (5)$ \\
			\hline
			average ranks &  1.83 & 3.00 & 4.00   & 1.67  & 4.33  \\
			\hline 
			H1 $\rightarrow$ H2 & $0.065 \pm 0.001 (2)$ & $ 0.061 \pm 0.000 (1) $ & $0.100 (5)$ & $0.076 (3)$ & $0.083 \pm 0.016 (4)$ \\
			\hline
			average ranks & 2.00   & 1.00  &  5.00   & 3.00   &  4.00  \\
			\hline
			U1 $\rightarrow$ U2 & $0.173 \pm 0.018 (3)$ & $0.093 \pm 0.000 (1)$ & $0.245 (4)$ & $0.246 (5)$ &  $0.114 \pm 0.012 (2)$ \\
			
			U2 $\rightarrow$ U1 & $0.232 \pm 0.005 (5)$ & $0.078 \pm 0.000 (1)$ & $0.218 (3)$ & $0.230 (4)$ &  $0.107 \pm 0.010 (2)$ \\
			\hline
			average ranks & 4.00   & 1.00  &  3.50   & 4.50   &  2.00  \\
			\hline
			V1 $\rightarrow$ V2 & $0.030 \pm 0.000 (2)$ & $0.030 \pm 0.000 (2)$ & $0.091 (4)$ & $0.002 (1)$ & $0.120 \pm 0.046 (5)$ \\
			
			V2 $\rightarrow$ V1 & $0.015 \pm 0.000 (1)$ & $0.038 \pm 0.008 (3)$ & $0.773 (5)$ & $0.015 (1)$ &  $0.070 \pm 0.032 (4)$ \\
			\hline
			average ranks & 1.50   & 2.50   & 4.50    & 1.00    &  4.50  \\
			\hline
		\end{tabular}
	\end{table*}
	
	\subsection{Experimental Setup}
	
	For the analogy-based methods, an important pre-processing step is the normalization of the attributes in the feature representation $\boldsymbol{x}=(x_1, \ldots , x_d)$, because these attributes are assumed to take values in $[0,1]$. To this end, we simply apply a linear rescaling 
	\[
	x_k'  \leftarrow \dfrac{x_k -\min_k}{ \max_k - \min_k } \, ,
	\]
	where $\min_k$ and $\max_k$ denote, respectively, the smallest and largest value of the $k$th feature in the data. This transformation is applied to the training data as well as the test data when a new query $Q$ is received. Since the data from a new query is normally sparse, it might be better to take the minimum and maximum over the entire data, training and test. Yet, this strategy is not recommendable in case the test data has a different distribution. As already said, analogical inference is especially interesting for transfer learning (and indeed, in our experiments, training and test data are sometimes from different subdomains). Therefore, we first conduct a Kolmogorov-Smirnov test \cite{kstest} to test whether the two parts of the data are drawn from the same distribution. In case the null hypothesis is rejected (at a significance level of $\alpha = 0.05$), normalization is conducted on the test data alone. Otherwise, the training data is additionally taken into account.
	
	We also apply a standard normalization for the other baseline methods (ERR, Ranking SVM and RankNet), transforming each real-valued feature by standardization:
	\[
	x  \leftarrow \dfrac{x-\mu}{\sigma} \, ,
	\]
	where $\mu$ and $\sigma$ denote the empirical mean and standard deviation, respectively. Like for the analogy-based methods, a hypothesis test is conducted to decide whether the test data should be normalized separately or together with the training data. 
	
	We fixed the cost parameter $C$ of SVM algorithms in an (internal) 2-fold cross validation (repeated 3 times) on the training data. The search for $C$ is guided by an algorithm\footnote{Publicly available as an R package: \url{http://cran.r-project.org/web/packages/svmpath}} proposed by \cite{Hastie2004}, which computes the entire regularization path for the two-class SVM classifier (i.e., all possible values of $C$ for which the solution changes), with a cost a small ($\sim$3) multiple of the cost of fitting a single model.
	The following RankNet parameters are adjusted using grid-search and internal cross-validation: the number of units in the hidden layer ($32,64,128,256$), the batch size ($8, 16, 32$), the optimizer learning rate ($0.001,0.01,0.1$). Since the data sets are relatively small, the network was restricted to a single hidden layer. 
	
	\subsection{Results}
	
	In our experiments, predictions were produced for certain data set $D_{test}$ of the data, using other parts $D_{train}$ as training data; an experiment of that kind is denoted by $D_{train} \rightarrow D_{test}$ that is considered for all possible combinations within each domain. The averaged ranking loss together with the standard deviation of the conducted experiments (repeated 20 times) are summarized in Table (\ref{tab:results}), where the numbers in parentheses indicate the rank of the achieved score in the respective problem. 
	Moreover, the table shows average ranks per problem domain. 
	
	As can be seen, the relative performance of the methods depends on the domain. In any case, our proposed approach is quite competitive in terms of predictive accuracy, and essentially on a par with able2rank and Ranking SVM, whereas ERR and RankNet show worse performance.
	
	\section{Conclusion and Future Work}
	
	This paper elaborates on the connection between kernel-based machine learning and analogical reasoning in the context of preference learning. Building on the observation that analogical proportions define a kind of similarity between the relation of pairs of objects, and that kernel functions can be interpreted in terms of similarity, we utilize generalized (fuzzy) equivalence relations as a bridging concept to show that a particular type of analogical proportion defines a valid kernel function. We introduce the analogy kernel and advocate a concrete kernel-based approach for the problem of object ranking. First experimental results on real-world data from various domains are quite promising and suggest that our approach is competitive to state-of-the-art methods for object ranking.

	By making analogical inference amenable to kernel methods, our paper depicts a broad spectrum of directions for future work. In particular, we plan to study kernel properties of other analogical proportions proposed in the literature (e.g., geometric proportions). 
	
	Besides, various extensions in the direction of kernel-based methods are conceivable and highly interesting from the point of view of analogical reasoning. This includes the use of kernel-based methods other than SVM, techniques such as multiple kernel learning, etc. Last but not least, other types of applications, whether in preference learning or beyond, are also of interest.  
	
	\bibliography{bibfile}
	\bibliographystyle{ieeetr}

\end{document}